\documentclass[runningheads]{llncs}
\usepackage[T1]{fontenc}
\usepackage{graphicx, amsmath, amssymb, amsfonts, caption, subcaption, multirow, overpic, textpos}
\usepackage{booktabs}
\usepackage{placeins}
\usepackage{bm}
\usepackage[misc]{ifsym} 
\usepackage[colorlinks=true]{hyperref}
\begin{document}
\title{OpenAL: An Efficient Deep Active Learning Framework for Open-Set Pathology Image Classification}
\titlerunning{OpenAL}


\author{
Linhao Qu$^*$\inst{1,2} \and
Yingfan Ma$^*$\inst{1,2} \and
Zhiwei Yang\inst{2,3} \and
Manning Wang$^{(\textrm{\Letter})}$\inst{1,2} \and
Zhijian Song$^{(\textrm{\Letter})}$\inst{1,2} }

\authorrunning{L. Qu \textit{et al.}}

%
%
\institute{Digital Medical Research Center, School of Basic Medical Science, Fudan University, Shanghai 200032, China \and 
Shanghai Key Lab of Medical Image Computing and Computer Assisted Intervention , Shanghai 200032, China \and 
Academy for Engineering \& Technology, Fudan University, Shanghai 200433, China.\\ \email{\{lhqu20, mnwang, zjsong\}@fudan.edu.cn}}
%


\renewcommand{\thefootnote}{}
\footnotetext{$^*$Linhao Qu and Yingfan Ma contributed equally.}
\maketitle              
\begin{abstract}
Active learning (AL) is an effective approach to select the most informative samples to label so as to reduce the annotation cost. 
Existing AL methods typically work under the closed-set assumption, i.e., all classes existing in the unlabeled sample pool need to be classified by the target model. 
However, in some practical clinical tasks, the unlabeled pool may contain not only the target classes that need to be fine-grainedly classified, but also non-target classes 
that are irrelevant to the clinical tasks. Existing AL methods cannot work well in this scenario because they tend to select a large number of non-target samples. In this paper, 
we formulate this scenario as an open-set AL problem and propose an efficient framework, OpenAL, to address the challenge of querying samples from an unlabeled pool with both 
target class and non-target class samples. Experiments on fine-grained classification of pathology images show that OpenAL can significantly improve the query quality of target 
class samples and achieve higher performance than current state-of-the-art AL methods. Code is available at https://github.com/miccaiif/OpenAL.

\end{abstract}
\section{Introduction}
Deep learning techniques have achieved unprecedented success in the field of medical image classification, but this is largely due to large amount of annotated data \cite{5,6,7}. 
However, obtaining large amounts of high-quality annotated data is usually expensive and time-consuming, especially in the field of pathology image processing \cite{6,7,100,101,102}. Therefore, 
a very important issue is how to obtain the highest model performance with a limited annotation budget.

\begin{figure*}[t]
    \centering
    \includegraphics[scale=0.43]{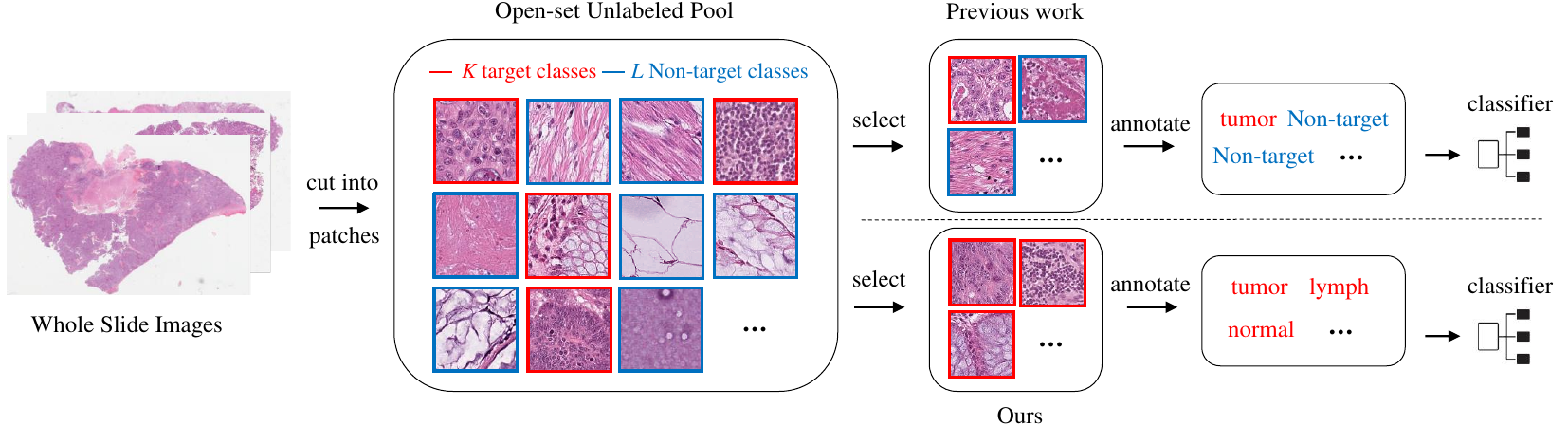}
    \caption{Description of the open-set AL scenario for pathology image classification. The unlabeled sample pool contains $K$ target categories (red-boxed images) and $L$ non-target categories 
    (blue-boxed images). Existing AL methods cannot accurately distinguish whether the samples are from the target classes or not, thus querying a large number of non-target samples and wasting the 
    annotation budget, while our method can accurately query samples from the target categories.}
    \label{figure1}
\end{figure*}

Active learning (AL) is an effective approach to address this issue from a data selection perspective, which selects the most informative samples from an unlabeled sample pool for 
experts to label and improves the performance of the trained model with reduced labeling cost \cite{9,4,11,12,13,14,15}. However, existing AL methods usually work under the closed-set assumption, i.e., 
all classes existing in the unlabeled sample pool need to be classified by the target model, which does not meet the needs of some real-world scenarios \cite{1}. Fig. \ref{figure1} shows an AL scenario 
for pathology image classification in an open world, which is very common in clinical practice. In this scenario, the Whole Slide Images (WSIs) are cut into many small patches that compose 
the unlabeled sample pool, where each patch may belong to tumor, lymph, normal tissue, fat, stroma, debris, background, and many other categories. However, it is not necessary to perform 
fine-grained annotation and classification for all categories in clinical applications. For example, in the cell classification task, only patches of tumor, lymphatic and normal cells need 
to be labeled and classified by the target model. Since the non-target patches are not necessary for training the classifier, labeling them would waste a large amount of budget. We call 
this scenario in which the unlabeled pool consists of both target class and non-target class samples open-set AL problem. Most existing AL algorithms can only work in the closed-set setting. 
Even worse, in the open-set setting, they even query more non-target samples because these samples tend to have greater uncertainty compared to the target class samples \cite{1}. Therefore, 
for real-world open-set pathology image classification scenarios, an AL method that can accurately query the most informative samples from the target classes is urgently needed.

Recently, Ning et al. \cite{1} proposed the first AL algorithm for open-set annotation in the field of natural images. They first trained a network to detect target class samples using a small number of 
initially labeled samples, and then modeled the maximum activation value (MAV) distribution of each sample using a Gaussian mixture model \cite{16} (GMM) to actively select the most deterministic target 
class samples for labeling. Although promising performance is achieved, their detection of target class samples is based on the activation layer values of the detection network which has limited accuracy 
and high uncertainty with small initial training samples.

In this paper, we propose a novel AL framework under an open-set scenario, and denote it as OpenAL, which cannot only query as many target class samples as possible but also query the most informative samples 
from the target classes. OpenAL adopts an iterative query paradigm and uses a two-stage sample selection strategy in each query. In the first stage, we do not rely on a detection network to select target 
class samples and instead, we propose a feature-based target sample selection strategy. Specifically, we first train a feature extractor using all samples in a self-supervised learning manner, and map all 
samples to the feature space. There are three types of samples in the feature space, the unlabeled samples, the target class samples labeled in previous iterations, and the non-target class samples queried in 
previous iterations but not being labeled. Then we select the unlabeled samples that are close to the target class samples and far from the non-target class samples to form a candidate set. In the second 
stage, we select the most informative samples from the candidate set by utilizing a model-based informative sample selection strategy. In this stage, we measure the uncertainty of all unlabeled samples in 
the candidate set using the classifier trained with the target class samples labeled in previous iterations, and select the samples with the highest model uncertainty as the final selected samples in this 
round of query. After the second stage, the queried samples are sent for annotation, which includes distinguishing target and non-target class samples and giving a fine-grained label to every target 
class sample. After that, we train the classifier again using all the fine-grained labeled target class samples. 

We conducted two experiments with different matching ratios (ratio of the number of target 
class samples to the total number of samples) on a public 9-class colorectal cancer pathology image dataset. The experimental results demonstrate that OpenAL can significantly improve the query quality of 
target class samples and obtain higher performance with equivalent labeling cost compared with the current state-of-the-art AL methods. To the best of our knowledge, this is the first open-set AL work in 
the field of pathology image analysis.

\section{Method}
We consider the AL task for pathology image classification in an open-set scenario. The unlabeled sample pool $P_U$ consists of $K$ classes of target samples and $L$ classes of non-target samples 
(usually, $K<L$). $N$ iterative queries are performed to query a fixed number of samples in each iteration, and the objective is to select as many target class samples as possible from $P_U$ in each query, 
while selecting as many informative samples as possible in the target class samples. Each queried sample is given to experts for labeling, and the experts will give fine-grained category labels for 
target class samples, while only giving a "non-target class samples" label for non-target class samples. 

\subsection{Framework Overview}
\begin{figure*}[t]
    \centering
    \includegraphics[scale=0.42]{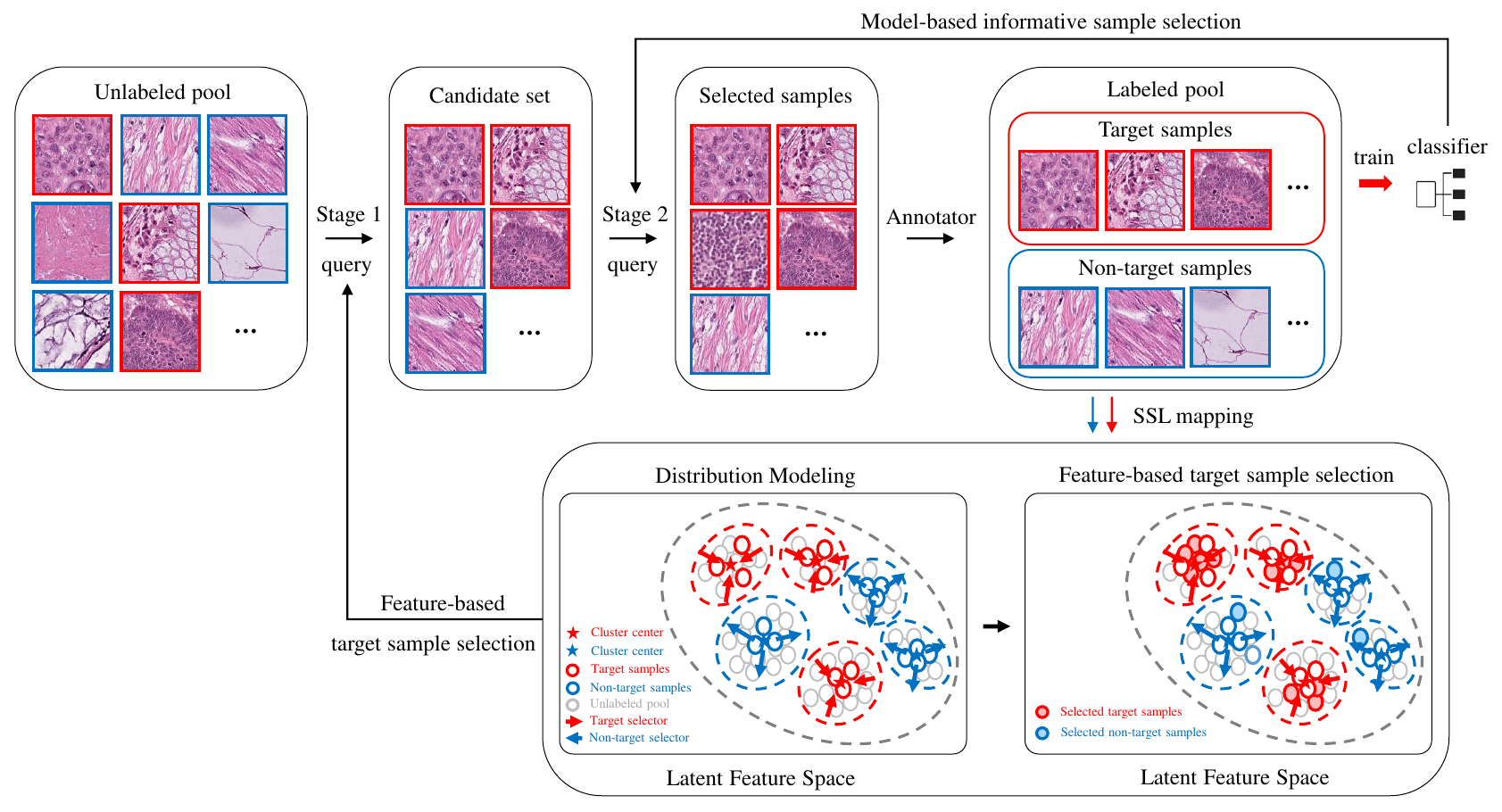}
    \caption{Workflow of OpenAL.}
    \label{figure2}
\end{figure*}

Fig. \ref{figure2} illustrates the workflow of the proposed method, OpenAL. OpenAL performs a total of $N$ iterative queries, and each query is divided into two stages. In Stage 1, OpenAL uses a feature-based target 
sample selection (FTSS) strategy to query the target class samples from the unlabeled sample pool to form a candidate set. Specifically, we first train a feature extractor with all samples by self-supervised 
learning, and map all samples to the feature space. Then we model the distribution of all unlabeled samples, all labeled target class samples from previous iterations, and all non-target class samples 
queried in previous iterations in the feature space, and select the unlabeled samples that are close to the target class samples and far from the non-target class samples. In Stage 2, OpenAL adopts a 
model-based informative sample selection (MISS) strategy. Specifically, we measure the uncertainty of all unlabeled samples in the candidate set using the classifier trained in the last iteration, and select 
the samples with the highest model uncertainty as the final selected samples, which are sent to experts for annotation. After obtaining new labeled samples, we train the classifier using all fine-grained 
labeled target class samples with cross-entropy as the loss function. The FTSS strategy is described in Section \ref{sec22}, and the MISS strategy is described in Section \ref{sec23}.

\subsection{Feature-based Target Sample Selection}
\label{sec22}

\textbf{Self-supervised Feature Representation.} First, we use all samples to train a feature extractor by self-supervised learning and map all samples to the latent feature space. Here, we adopt DINO \cite{17,19} as the self-supervised network because of its outstanding performance.

\textbf{Sample Scoring and Selection in the Feature Space.} Then we define a scoring function on the base of the distribution of unlabeled samples, labeled target class samples and non-target class samples queried in previous iterations. Every unlabeled sample in the current iteration is given a score, and a smaller score indicates that the sample is more likely to come from the target classes. The scoring function is defined in Equation \ref{eq1}.

\begin{equation}
s_i=s_{t_i}-s_{w_i}
    \label{eq1}
\end{equation}
where $s_i$ denotes the score of the unlabeled sample $x_i^U$ . $s_{t_i}$ measures the distance between $x_i^U$ and the distribution of features derived from all the labeled target class samples. 
The smaller $s_{t_i}$ is, the closer $x_i^U$ is to the known sample distribution of the target classes, and the more likely $x_i^U$ is from a target class. Similarly, $s_{w_i}$ measures the distance 
between $x_i^U$ and the distribution of features derived from all the queried non-target class samples. The smaller $s_{w_i}$ is, the closer $x_i^U$ is from the known distribution of non-target class 
samples, and the less likely $x_i^U$ is from the target class. After scoring all the unlabeled samples, we select the top $\varepsilon$\% samples with the smallest scores to form the candidate set. 
In this paper, we empirically take twice the current iterative labeling budget (number of samples submitted to experts for labeling) as the sample number of the candidate set. Below, we give 
the definitions of $s_{t_i}$ and $s_{w_i}$.

\textbf{Distance-based Feature Distribution Modeling.} We propose a category and Mahalanobis distance-based feature distribution modeling approach for calculating $s_{t_i}$ and $s_{w_i}$. 
The definitions of these two values are slightly different, and we first present the calculation of $s_{t_i}$, followed by that of $s_{w_i}$. 

For all labeled target class samples from previous iterations, their fine-grained labels are known, so we represent these samples as different clusters in the feature space according to their true class 
labels, where a cluster is denoted as $C_t^L (t=1, …, K)$. Next, we calculate the score $s_{t_i}$ for $z_i^U$ using the Mahalanobis distance (MD) according to Equation \ref{eq2}.
MD is widely used to measure the distance between a point and a distribution because it takes into account the mean and variance of the distribution, which is very suitable for our scenario.
\begin{equation}
    s_{t_i }=\operatorname{Nom}\left(\min _t\left(D\left(z_i^U, C_t^L\right)\right)\right)=\operatorname{Nom}\left(\min _t\left(z_i^U-\mu_t\right)^T \Sigma_t^{-1}\left(z_i^U-\mu_t\right)\right) \label{eq2}
    \end{equation}

\begin{equation}
\operatorname{Nom}(X)=\frac{X-X_{\min }}{X_{\max }-X_{\min }} \label{eq3}
\end{equation}
where $D(\cdot)$ denotes the MD function, $\mu_t$ and $\Sigma_t$ are the mean and covariance of the samples in the target class $t$, and $\operatorname{Nom}(\cdot)$ is the normalization function. 
It can be seen that $s_{t_i }$ is essentially the minimum distance of the unlabeled sample $x_i^U$ to each target class cluster.

For all the queried non-target class samples from previous iterations, since they do not have fine-grained labels, we first use the K-means algorithm to cluster their features into $w$ classes, where a cluster is denoted as $C_w^L$ $(w=1, …,W)$. $W$ is set to 9 in this paper. Next, we calculate the score $s_{w_i }$ for $z_i^U$ using the MD according to Equation \ref{eq4}.

\begin{equation}
    s_{w_i}=\operatorname{Nom}\left(\min _w\left(D\left(z_i^U, C_w^L\right)\right)\right)=\operatorname{Nom}\left(\min _w\left(z_i^U-\mu_w\right)^T \Sigma_t^{-1}\left(z_i^U-\mu_w\right)\right) \label{eq4}
\end{equation}
where $\mu_w$ and $\mathrm{\Sigma}_w$ are the mean and covariance of the non-target class sample features in the $w$th cluster. It can be seen that $s_{w_i}$ is essentially 
the minimum distance of $z_i^U$ to each cluster of known non-target class samples.

The within-cluster selection and dynamic cluster changes between rounds significantly enhance the diversity of the selected samples and reduce redundancy.

\subsection{Model-based Informative Sample Selection}
\label{sec23}
To select the most informative samples from the candidate set, we utilize the model-based informative sample selection strategy in Stage 2. We measure the uncertainty of 
all unlabeled samples in the candidate set using the classifier trained in the last iteration and select the samples with the highest model uncertainty as the final selected samples. The entropy 
of the model output is a simple and effective way to measure sample uncertainty \cite{2,3}. Therefore, we calculate the entropy of the model for the samples in the candidate set and select 
50\% of them with the highest entropy as the final samples in the current iteration.

\section{Experiments}
\subsection{Dataset, Settings, Metrics and Competitors}
To validate the effectiveness of OpenAL, we conducted two experiments with different matching ratios (the ratio of the number of samples in the target class to the total number of samples) 
on a 9-class public colorectal cancer pathology image classification dataset (NCT-CRC-HE-100K) \cite{18}. The dataset contains a total of 100,000 patches of pathology images with fine-grained labeling, 
with nine categories including Adipose (ADI 10\%), background (BACK 11\%), debris (DEB 11\%), lymphocytes (LYM 12\%), mucus (MUC 9\%), smooth muscle (MUS 14\%), normal colon mucosa (NORM 9\%), 
cancer-associated stroma (STR 10\%), and colorectal adenocarcinoma epithelium (TUM, 14\%). To construct the open-set datasets, we selected three classes, TUM, LYM and NORM, as the target 
classes and the remaining classes as the non-target classes. We selected these target classes to simulate a possible scenario for pathological cell classification in clinical practice. Technically, target classes can be randomly chosen.
In the two experiments, we set the matching ratio to 33\% (3 target classes, 6 non-target classes), and 42\% (3 target classes, 4 
non-target classes), respectively.

\textbf{Metrics.} Following \cite{1}, we use three metrics, precision, recall and accuracy to compare the performance of each AL method. We use precision and recall to measure the performance of 
different methods in target class sample selection. As defined in Equation \ref{eq5}, precision is the proportion of the target class samples among the total samples queried in each query and 
recall is the ratio of the number of the queried target class samples to the number of all the target class samples in the unlabeled sample pool.

\begin{equation}
\text { precision}_m=\frac{k_m}{k_m+l_m}, \operatorname{recall}_m=\frac{\sum_{j=0}^m k_m}{n_{\text {target }}}  \label{eq5}
\end{equation}
where $k_m$ denotes the number of target class samples queried in the $m$th query, $l_m$ denotes the number of non-target class samples queried in the $m$th query, and $n_{target}$ denotes the 
number of target class samples in the original unlabeled sample pool. Obviously, the higher the precision and recall are, the more target class samples are queried, and the more effective the 
trained target class classifier will be. We measure the final performance of each AL method using the accuracy of the final classifier on the test set of target class samples.

\textbf{Competitors.} We compare the proposed OpenAL to random sampling and five AL methods, LfOSA \cite{1}, Uncertainty \cite{2,3}, Certainty \cite{2,3}, Coreset \cite{4} and RA \cite{5}, of 
which only LfOSA \cite{1} is designed for open-set AL. For all AL methods, we randomly selected 1\% of the samples to label and used them as the initial labeled set for model initialization. It is 
worth noting that the initial labeled samples contain target class samples as well as non-target class samples, but the non-target class samples are not fine-grained labeled. After each query round, 
we train a ResNet18 model of 100 epochs, using SGD as the optimizer with momentum of 0.9, weight decay of 5e-4, initial learning rate of 0.01, and batchsize of 128. The annotation budget for each 
query is 5\% of all samples, and the length of the candidate set is twice the budget for each query. For each method, we ran four experiments and recorded the average results for four randomly selected seeds.

\begin{figure*}[t!]
    \centering
    \includegraphics[scale=0.41]{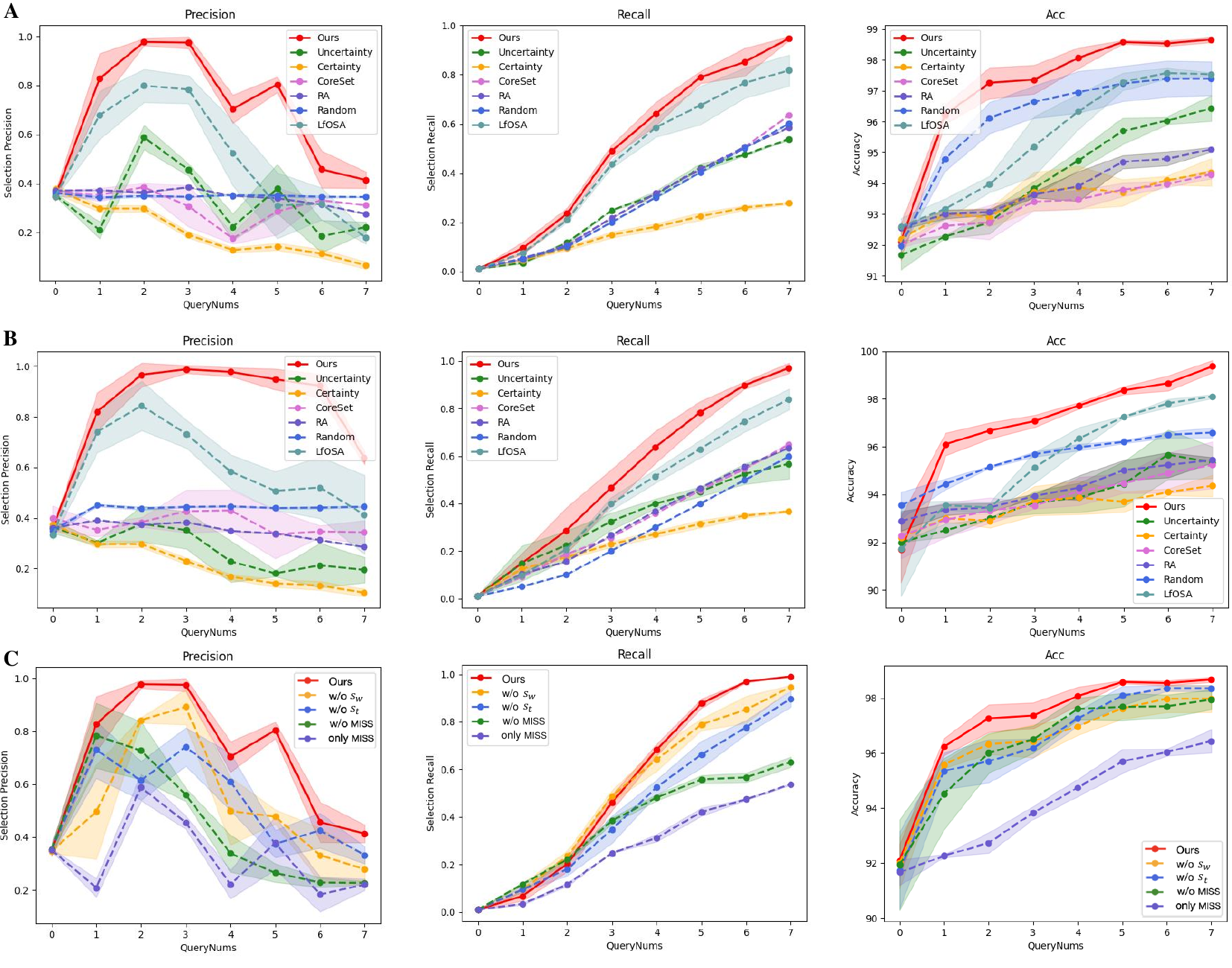}
    \caption{A. Selection and model performance results under a 33\% matching ratio. B. Selection and model performance results under a 42\% matching ratio. C. Ablation Study of OpenAL under a 33\% matching ratio.}
    \label{figure3}
\end{figure*}

\begin{figure*}[t!]
    \centering
    \includegraphics[scale=0.61]{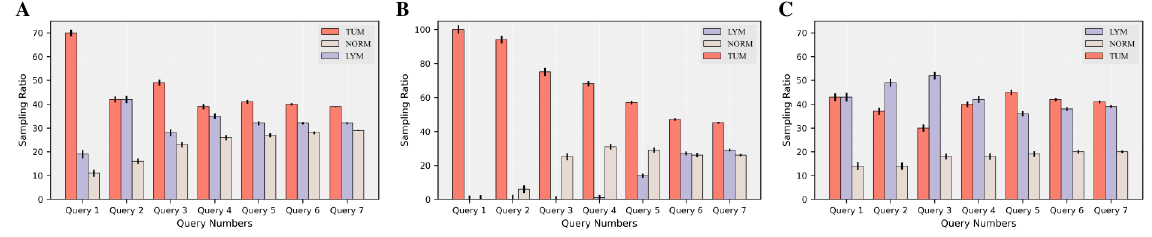}
    \caption{A. Cumulative sampling ratios of our OpenAL for the target classes LYM, NORM, and TUM across QueryNums 1-7 on the original dataset (under 33\% matching ratio). 
    B. Cumulative sampling ratios of LfOSA on the original dataset (under 33\% matching ratio). 
    C. Cumulative sampling ratios of our OpenAL on a newly-constructed more imbalanced setting for the target classes LYM (6000 samples), NORM (3000 samples), and TUM (9000 samples).}
    \label{figure4}
\end{figure*}

\subsection{Performance Comparison}
Fig. \ref{figure3} A and B show the precision, recall and model accuracy of all comparing methods at 33\% and 42\% matching ratios, respectively. It can be seen that OpenAL outperforms the other 
methods in almost all metrics and all query numbers regardless of the matching ratio. Particularly, OpenAL significantly outperforms LfOSA \cite{1}, which is specifically designed for open-set AL. 
The inferior performance of the AL methods based on the closed-set assumption is due to the fact that they are unable to accurately identify more target class samples, thus wasting a large amount of 
annotation budget. Although LfOSA \cite{1} utilizes a dedicated network for target class sample detection, the performance of the detection network is not stable when the number of training samples is 
small, thus limiting its performance. In contrast, our method uses a novel feature-based target sample selection strategy and achieves the best performance.

Upon analysis, our OpenAL is capable of effectively maintaining the balance of sample numbers across different classes during active learning. 
We visualize the cumulative sampling ratios of OpenAL for the target classes in each round on the original dataset with a 33\% matching ratio, as shown in Fig. \ref{figure4} A. 
Additionally, we visualize the cumulative sampling ratios of the LfOSA method on the same setting in Fig. \ref{figure4} B. It can be observed that in the first 4 rounds, 
LYM samples are either not selected or selected very few times. This severe sample imbalance weakens the performance of LfOSA compared to random selection initially. 
Conversely, our method selects target class samples with a more balanced distribution. Furthermore, we constructed a more imbalanced setting for the target classes LYM 
(6000 samples), NORM (3000 samples), and TUM (9000 samples), yet the cumulative sampling ratios of our method for these three target classes remain fairly balanced, as shown in Fig. \ref{figure4} C.

\subsection{Ablation Study}
To further validate the effectiveness of each component of OpenAL, we conducted an ablation test at a matching ratio of 33\%. Fig. \ref{figure3} C shows the results, 
where w/o $s_w$ indicates that the distance score of non-target class samples is not used in the scoring of Feature-based Target Sample Selection (FTSS), w/o $s_t$ indicates that the 
distance score of target class samples is not used, w/o MISS means no Model-based Informative Sample Selection is used, i.e., the length of the candidate set is directly set to the 
annotation budget in each query, and only MISS means no FTSS strategy is used, but only uncertainty is used to select samples.

It can be seen that the distance modeling of both the target class samples and the non-target class samples is essential in the FTSS strategy, and missing either one results in a 
decrease in performance. Although the MISS strategy does not significantly facilitate the selection of target class samples, it can effectively help select the most informative samples 
among the samples in the candidate set, thus further improving the model performance with a limited labeling budget. In contrast, when the samples are selected based on uncertainty alone, 
the performance decreases significantly due to the inability to accurately select the target class samples. The above experiments demonstrate the effectiveness of each component of OpenAL.

\section{Conclusion}
In this paper, we present a new open-set scenario of active learning for pathology image classification, which is more practical in real-world applications. 
We propose a novel AL framework for this open-set scenario, OpenAL, which addresses the challenge of accurately querying the most informative target class samples 
in an unlabeled sample pool containing a large number of non-target samples. OpenAL significantly outperforms state-of-the-art AL methods on real pathology image classification tasks. 
More importantly, in clinical applications, on one hand, OpenAL can be used to query informative target class samples for experts to label, thus enabling better training of 
target class classifiers under limited budgets. On the other hand, when applying the classifier for future testing, it is also possible to use the feature-based target sample 
selection strategy in the OpenAL framework to achieve an open-set classifier. Therefore, this framework can be applied to both datasets containing only target class samples and 
datasets also containing a large number of non-target class samples during testing.

\section*{Acknowledgments}
This work was supported by National Natural Science Foundation of China under Grant 82072021.

%
%
%
\bibliographystyle{splncs04}
\bibliography{OpenAL}
\end{document}